\documentclass{article}

\usepackage{arxiv}

\usepackage[utf8]{inputenc} % allow utf-8 input
\usepackage[T1]{fontenc}    % use 8-bit T1 fonts
\usepackage{hyperref}       % hyperlinks
\usepackage{url}            % simple URL typesetting
\usepackage{booktabs}       % professional-quality tables
\usepackage{amsfonts}       % blackboard math symbols
\usepackage{nicefrac}       % compact symbols for 1/2, etc.
\usepackage{microtype}      % microtypography
\usepackage{lipsum}
\usepackage{graphicx}
\usepackage{indentfirst}
\graphicspath{ {./images/} }
\usepackage[table,xcdraw]{xcolor}
\usepackage{enumerate}
\usepackage[shortlabels]{enumitem}

\usepackage{authblk}

\usepackage{wrapfig}
\usepackage{graphicx}
\usepackage{picinpar}

\makeatletter
\long\def\figwindownonum[#1,#2,#3,#4] {% \begin{figwindownonum}
  \begin{window}[#1,#2,{#3},{\centering#4\par}] }
\def\endfigwindownonum{\end{window}}% \end{figwindownonum}
\makeatother

\title{The SkatingVerse Workshop \& Challenge: \\
Methods and Results}

\author{Jian Zhao$^{1,2}$, Lei Jin$^{3}$, Jianshu Li$^{4}$, Zheng Zhu$^{5}$, Yinglei Teng$^{3}$, Jiaojiao Zhao$^{6}$, 
Sadaf Gulshad$^{6}$, Zheng Wang$^{7}$, Bo Zhao$^{8}$, Xiangbo Shu$^{9}$, Yunchao Wei$^{10}$, Xuecheng Nie$^{11}$, Xiaojie Jin$^{12}$, Xiaodan Liang$^{13}$, , Shin'ichi Satoh$^{14}$, Yandong Guo$^{15}$, Cewu Lu$^{16}$, Junliang Xing$^{17}$, Jane Shen Shengmei$^{18}$ \\
{\small $^{1}$EVOL Lab, Institute of AI (TeleAI), China Telecom, $^{2}$Northwestern Polytechnical University, $^{3}$Beijing University of Posts and Telecommunications, $^{4}$Ant Group} 
{\small $^{5}$GigaAI, $^{6}$University of Amsterdam, $^{7}$Wuhan University,} 
{\small $^{8}$BMO AI, $^{9}$Nanjing University of Science and Technology, $^{10}$Beijing Jiaotong University, $^{11}$MT lab,}
{\small $^{12}$Bytedance Inc., $^{13}$Sun Yat-sen University, $^{14}$The University of Tokyo,} 
{\small $^{15}$$\mathrm{AI}^2$ Roboitcs, $^{16}$Shanghai Jiao Tong University, $^{17}$Tsinghua University, $^{18}$Mikomiko} 
}

\begin{document}
\maketitle
\begin{abstract}
The SkatingVerse Workshop \& Challenge aims to encourage research in developing novel and accurate methods for human action understanding. The SkatingVerse dataset used for the SkatingVerse Challenge has been publicly released. There are two subsets in the dataset, $i.e.$, the training subset and testing subset. The training subsets consists of 19,993 RGB video sequences, and the testing subsets consists of 8,586 RGB video sequences. Around 10 participating teams from the globe competed in the SkatingVerse Challenge. In this paper, we provide a brief summary of the SkatingVerse Workshop \& Challenge including brief introductions to the top three methods. The submission leaderboard will be reopened for researchers that are interested in the human action understanding challenge. The benchmark dataset and other information can be found at: https://skatingverse.github.io/.
\end{abstract}

% keywords can be removed

\section{Introduction}

Human action understanding (HAU) is an essential topic in computer vision. Typically, HAU aims to locate, classify, and assess the human actions from a given video. Many tasks, e.g., action recognition, action segmentation, action localization, and action assessment, belong to the research scope of HAU. However, in a real-world scenario, it is typically necessary not only to segment each fine-grained action but also to assess their quality. Existing tasks and corresponding methods alone cannot meet this requirement. To this end, we construct a dataset consisting of 1,687 origal continuous videos in figure skating competition. We encourage participants to develop intelligent computer vision algorithms that can segment and assess each single action in an original and continuous competition video. Particularly, algorithms that can distinguish exactly similar actions and give a score based on action competition are highly expected. We hope this challenge can further promote the machine's perceptual cognition, which is a big step in real artificial intelligence.

This workshop will bring together academic and industrial experts in the field of HAU to discuss the techniques and applications of HAU. Participants are invited to submit their original contributions, surveys, and case studies that address the human actions perception and understanding issues.

\begin{figure}[h] 
\vspace{-2mm}
\begin{center}
\includegraphics[width=0.95\linewidth]{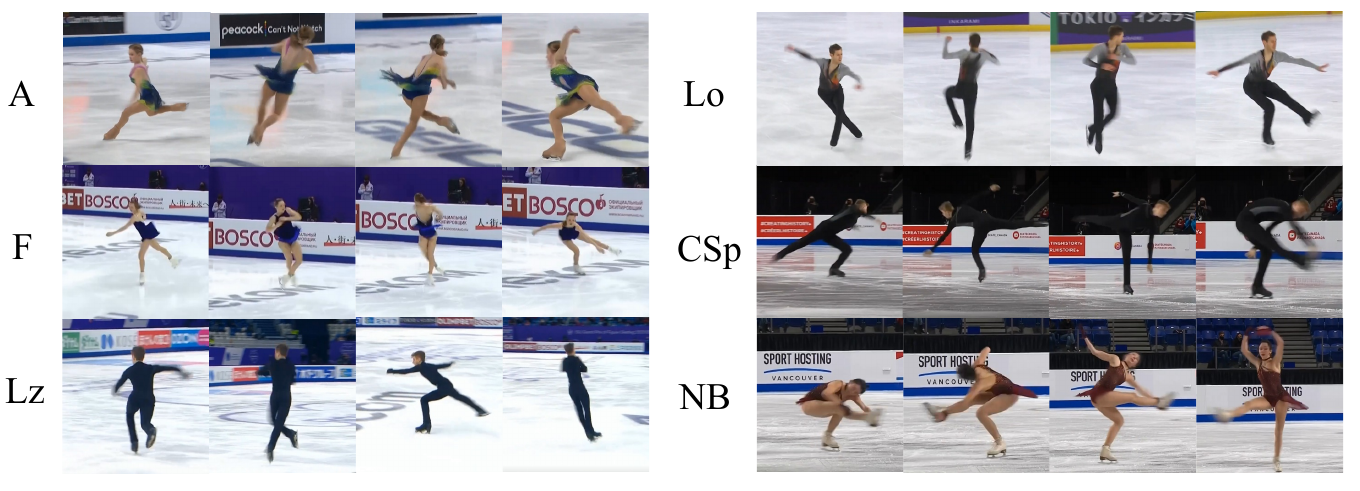}
\end{center}
   \caption{Examples of figure skating actions: A (Axel); Lo (Loop); F (Flip); CSp (Camel Spin); Lz (Lutz); NB (No Basic).
}
\label{fig:1}
\end{figure}

\section{The Skating Challenge}
\subsection{Dataset}
This challenge only involves human action recognition task. We preprocess the original figure skating videos to construct the human action recognition dataset. There are two subsets in the dataset, $i.e.$, the training subset and testing subset. The former subset consists of 19993 videos which used for training algorithm, and the latter subset consists of 8586 videos which used for evaluating the methods. We only provide annotation files for the training subset. We provide semantic annotations hierarchically in action labels. Specifically, we have two levels of categorical labels, namely set and element. In the set level, there are 11 action classes, including 6 jumps Toe loop (T), Loop (Lo), Axel (A), Lutz (Lz), Salchow (S), Flip (F), 4 spins Camel spin (CSp), Sit spin (SSp), Upright spin (USp), No Basic (NB) and 1 sequence (Null). While in the fine-grained element level, the actions are further divided into 28 categories according to the number of turns for jump actions.

\subsection{Metric}
We use `Top1 Acc’ and `Mean Acc’ metrics to evaluate the performance of algorithms in SkatingVerse Challenge. `Top1 Acc’ represents the proportion of successful predicted samples. For the total number of samples $N$ and the number of correct predicted samples $M$, `Top1 Acc’ refer to $\frac{M}{N}$. `Mean Acc’ denotes the average accuracy of all categories. For $K$ categories, the number of samples of $i^{th}$ class is $N_i$, the number of correct predicted samples of $i^{th}$ class is $M_i$, and the `Mean Acc’ can be calculated as follow:
\begin{equation}
    Mean\_Acc = \frac{1}{K} \sum_{i=1}^{K} \frac{M_i}{N_i}
\end{equation}

\section{Result and Method}
The 2nd Anti-UAV challenge was held between May 12, 2021 and July 10, 2021.The results of the 2nd Anti-UAV challenge are shown in Table 1. Around 24 teams submitted their final results in this challenge. In this section,we will briefly introduce the methodologies of the top 3 submissions.

\begin{table}[]
\centering
\caption{Challenge results}
\label{tab1}
\begin{tabular}{
>{\columncolor[HTML]{FFFFFF}}c 
>{\columncolor[HTML]{FFFFFF}}c 
>{\columncolor[HTML]{FFFFFF}}c }
\hline
{\color[HTML]{333333} \textbf{Rank}} & {\color[HTML]{333333} \textbf{User Name}}             & {\color[HTML]{333333} \textbf{Tracking Accuracy}} \\ \hline
{\color[HTML]{333333} 1}             & {\color[HTML]{333333} YuanziFu}             & {\color[HTML]{333333} 0.95727}                     \\
{\color[HTML]{333333} 2}           & {\color[HTML]{333333}Alex\_yang0828}                   & {\color[HTML]{333333} 0.95020}                     \\
{\color[HTML]{333333} 3}             & {\color[HTML]{333333} RunqingCMsS}                       & {\color[HTML]{333333} 0.88518}                     \\
{\color[HTML]{333333} 4}             & {\color[HTML]{333333} FengshengQiao}                 & {\color[HTML]{333333}0.86194}                     \\
{\color[HTML]{333333} 5}             & {\color[HTML]{333333} zhangheng}             & {\color[HTML]{333333} 0.85923}                     \\
{\color[HTML]{333333} 6}             & {\color[HTML]{333333} RungingZhang}                       & {\color[HTML]{333333} 0.82925}                     \\
{\color[HTML]{333333} 7}             & {\color[HTML]{333333} lizhexyz}                    & {\color[HTML]{333333} 0.79814}                     \\
{\color[HTML]{333333} 8}             & {\color[HTML]{333333} GuangzhaoDai1}                & {\color[HTML]{333333} 0.57886}                     \\
{\color[HTML]{333333} 9}             & {\color[HTML]{333333} cqbu}                   & {\color[HTML]{333333} 0.56845}                     \\
{\color[HTML]{333333} 10}            & {\color[HTML]{333333} inferno} & {\color[HTML]{333333} 0.41260}                     \\
\hline
\end{tabular}
\end{table}

\subsection{Team DeepGlint}
\textbf{Tao Sun, Yuanzi Fu, Kaicheng Yang, Jian Wu, Ziyong Feng.} (Beijing DeepGlint)

The authors propose a method that involves several steps. To begin, they leverage the DINO framework\cite{zhang2022dino} to extract the Region of Interest (ROI) and perform precise cropping of the raw video footage. Subsequently, they employ three distinct models, namely Unmasked Teacher\cite{li2023unmasked}, UniformerV2\cite{li2022uniformerv2}, and InfoGCN\cite{chi2022infogcn}, to capture different aspects of the data. By ensembling the prediction results based on logits,the solution attains an impressive leaderboard score of 95.73.

\textbf{Dataset Pre-processing.} In order to enhance the model’s attention toward figure skating actions, the authors performed human body detection on the original video frames. To accomplish this, they utilized FFmpeg to extract video frames and employed the DINO framework \cite{zhang2022dino} for extracting bounding boxes corresponding to human detections in each frame. These individual bounding box results from all frames within a video were then consolidated to generate the ultimate detection box for that specific video. Finally, using FFmpeg, they crop the raw video clips based on the combined bounding box information obtained from the human detection process.

\textbf{Model Structure.} The authors employ the DINO model \cite{zhang2022dino} to extract precise human detection bounding boxes, facilitating the generation of cropped frames. Subsequently, they conduct fine-tuning on two widely-used general video pretraining models, namely Unmasked Teacher \cite{li2023unmasked} and UniformerV2 \cite{li2022uniformerv2}. Furthermore, they leverage ViTPose \cite{xu2022vitpose} to extract human skeleton sequences, which are then utilized to train the InfoGCN model \cite{chi2022infogcn} for accurate action predictions.  

1) Unmasked Teacher: Unmasked Teacher(UMT) \cite{li2023unmasked} is a two-stage training-efficient pretraining framework that enhances temporal-sensitive video foundation models by incorporating the advantages of previous approaches. In Stage 1, the UMT exclusively utilizes video data for masked video modeling, resulting in a model that excels at video-only tasks. In Stage 2, the UMT leverages public vision-language data for multi-modality learning, enabling the model to handle complex video-language tasks such as video-text retrieval and video question answering.
The authors initially fine-tune the pre-trained UMT-L16 model (which is pre-trained and fine-tuned on Kinetices710 with 8 × 2242 input images) for 50 epochs using 16 × 2242 input images. Subsequently, based on the obtained fine-tuned model weights, they perform weight interpolation and further fine-tune the model for 10 additional epochs. This fine-tuning is conducted separately using both 16×4482 images and 32 × 2242 images. To obtain the final prediction result of UMT, they aggregate the predicted probabilities from the three models and apply the softmax function.  

2) UniformerV2: UniformerV2 \cite{li2022uniformerv2} is a versatile approach for building a robust collection of video networks. It combines the image-pretrained Vision Transformers (ViTs) with efficient video designs from UniFormer \cite{li2023uniformer}. The key innovation in UniformerV2 lies in the inclusion of novel local and global relation aggregators. These aggregators seamlessly integrate the strengths of both ViTs and UniFormer, achieving a desirable balance between accuracy and computational efficiency.In this work, the authors conduct fine-tuning on the UniFormerV2L14 model(use the CLIP \cite{radford2021learning} model weight pre-trained on LAION400M \cite{schuhmann2021laion}) for 30 epochs, utilizing 32 × 2242 input images.  

3) ViTPose \& InfoGCN: To obtain additional skatingrelated information from human skeleton sequences, the authors first use ViTPose \cite{xu2022vitpose} to extract the human skeleton sequence for each cropped frame. ViTPose utilizes plain and nonhierarchical vision transformers as backbones to extract features for individual person instances. It also incorporates a lightweight decoder for efficient pose estimation. ViTPose offers great flexibility in terms of attention type, input resolution, pre-training, fine-tuning strategies, and the ability to handle multiple pose tasks.Following the extraction of the human skeleton sequence for each cropped frame using ViTPose, they obtain a sequence of length 320 for each video. To accurately predict the categories of figure skating actions, they train the InfoGCN \cite{chi2022infogcn} model from scratch over the course of 300 epochs. This training process enables the model to effectively analyze and classify the various figure skating actions depicted in the videos.  

\textbf{Model Ensemble.} To further improve the leaderboard score, we adopt two simple ensemble strategies. The first approach involves voting among all the model predictions, with UniformerV2’s prediction being selected in case of a tie. The second approach involves weighted aggregation of the predictions, where the weights for UMT, UniformerV2, and InfoGCN are determined using the softmax function with weights [94.5, 95.0, 92.0], respectively.
   
\textbf{Ablation study.} The authors do ablation experiments to verify the impact of each algorithm, and they provided six sets of results, described below:

\begin{enumerate}[-,nosep]
    \item Unmasked Teacher  -> 94.55 
    \item UniformerV2 -> 95.02
    \item InfoGCN -> 92.03 
    \item Model Voter -> 95.67
    \item Model Weighted Summation -> 95.73
\end{enumerate}

\textbf{Main contribution.} Authors proposed to use DINO detection algorithm to remove the influence of irrelevant scenes and only focus on the action itself, which greatly improved the effect of the algorithm. Meanwhile, multiple algorithms were integrated to further improve the accuracy.

\subsection{Team CMSS-CXZX}
\textbf{Dunbo Cai, Zhiguo Huang, Liqiang Xu, Fengyun Zhuang, Wenjie Qin, 
 Feng Yang.} (China Mobile (Suzhou) Software Technology Co., Ltd.)
 
\textbf{Dataset Pre-processing.} The test set labels of the data set participating in the competition are not publicly available, therefore the authors opt to partition the training set into ten parts, with nine for training and one for testing and validation purposes. Additionally, the video labels are processed according to the K400 standard format\cite{kay2017kinetics}, facilitating ease of reading and training for the model.

\textbf{Base model.} Authors meticulously curated the current state-of-the-art open source action recognition algorithms, specifically selecting the VideoMae2 model\cite{wang2023videomae} as our foundational framework and opting for fine-tuning task.  The training weight was derived from distilling the VIT-base model\cite{dosovitskiy2020image}, while adhering to the K400 dataset structure in line with general training practices.

According to the inference results, authors analyze the accuracy and quantity of each category and observe that certain categories exhibit poor accuracy due to their limited representation in the dataset. Consequently, the model fail to acquire sufficient information about these categories, resulting in lower accuracy. Additionally, they identify a class imbalance issue within the data distribution of our dataset. To address these challenges, they implemented several solutions: 

   1) Augmenting samples for underrepresented classes during training sessions with an aim to enhance learning on minority categories.
   
   2) Employing  Label-Smoothing loss\cite{szegedy2016rethinking} as a replacement for regular cross-entropy loss to tackle unbalanced data distribution.
   
   3) Utilizing Focal loss\cite{lin2017focal} instead of cross-entropy loss.
   
   4) Based on the aforementioned outcomes, it became evident that class imbalance significantly impacted final scores. Therefore, they introduce weight multiplication for each class's loss based on its number of categories; larger numbers received smaller weights while smaller numbers were assigned larger weights. This punitive mechanism facilitated improve learning for fewer classes.
   
   5) Combining all these approaches together, they adopt a combination of Label-Smoothing loss\cite{szegedy2016rethinking} and distribution weight method.

\textbf{Model Ensemble.} Authors also train MVD \cite{wang2023masked}and UniformerV2\cite{li2022uniformerv2} for model ensemble. For hard ensemble, the voting base used is VideoMAE2\cite{wang2023videomae}, which exhibits the highest accuracy in the test set. For soft ensemble, the softmax category probabilities of videos corresponding to the three models are aggregated and assigned weights [0.5, 0.3, 0.2] respectively to obtain the final prediction result category.

\textbf{Ablation study.} 
\begin{enumerate}[-,nosep]
    \item VideoMae2  -> 76.6
    \item VideoMae2 + Eval -> 82.4
    \item VideoMae2 + Smoothlabel -> 85.3
    \item VideoMae2 + Focal loss -> 84.9
    \item VideoMae2 + Weight  -> 86.5
    \item MVD -> 84.6
    \item UniformerV2 -> 85.9
    \item Model Voter -> 87.9
    \item Model Weighted Summation  -> 88.5
\end{enumerate}
% SuperDiMP + AlphaRef  -> 60.6 
% SuperDiMP + AlphaRef + LTMU -> 64.0
% Stark -> 60.6 
% Stark + AlphaRef -> 62.1
% Stark + AlphaRef + LTMU -> 63.1
% multi-tracker voting and fusion -> 67.5
% +motion enhancement -> 68.7

\textbf{Main contribution.}
The authors provide technical solutions of various excellent action recognition algorithms on figure skating data sets, and optimizes and improves them according to the characteristics of specific data sets.

\subsection{Team ELK}
\textbf{Fengsheng Qiao.} (Chengdu University of Technology)

The authors use TPN algorithm\cite{yang2020temporal} for action recognition, train 150,400,600  epoch respectively, and vote the best weights of the three times for inference. Temporal Pyramid Network \cite{yang2020temporal} is a feature-level framework that revolutionizes the approach to action recognition in videos.   By introducing a temporal pyramid structure at the feature level, TPN efficiently captures the visual tempos of different actions, allowing the model to recognize actions with varying speeds and durations.   Unlike previous methods that rely on expensive multi-branch networks and raw video sampling, TPN seamlessly integrates into 2D or 3D backbone networks, offering a plug-and-play solution.   This flexibility allows TPN to enhance a wide range of existing action recognition models, as demonstrated by its consistent improvements over challenging baselines on multiple datasets.   Further analysis reveals that TPN particularly excels at recognizing action classes with large variations in visual tempos, validating its effectiveness and robustness.   In essence, TPN represents a novel approach to temporal modeling in action recognition, offering an efficient and powerful solution for capturing the dynamics and temporal scales of actions in videos.

% \textbf{Ablation study.} 
% \begin{enumerate}[-,nosep]
%     \item superDiMP  -> 56.3 
%     \item Multi-scale superDiMP -> 61.1
%     \item Multi-scale superDiMP + SiamPRN++ -> 62.7 
%     \item Multi-scale superDiMP + TansT + SiamRPN++ -> 64.0
%     \item Multi-scale SuperDiMP + TansT+ SiamRPN++ + Re-detection -> 67.9
% \end{enumerate}

\section{Conclusions}
Human action understanding is a fundamental yet challenging hotspot in computer vision. We held the 1 SkatingVerse Challenge to encourage researchers from the fields of human action understanding and other disciplines to present their progress, communication and novel ideas that will potentially shape the future of the HAU area. Approximately 10 teams around the globe participated in this competition, in which top-3 leading teams, together with their methods, are briefly introduced in this paper. Our workshop takes a different perspective, making figure skating as recognition target, and provides a fine-grained large-scale dataset to promote deep network learning for HAU. Thus, our workshop will bridge the needs of industry and research in academia, and may accelerate the process of these computer vision technologies being used in real applications.
Human action understanding is a fundamental yet challenging hotspot in computer vision. We held the 1 SkatingVerse Challenge to encourage researchers from the fields of human action understanding and other disciplines to present their progress, communication and novel ideas that will potentially shape the future of the HAU area. Approximately 10 teams around the globe participated in this competition, in which top-3 leading teams, together with their methods, are briefly introduced in this paper. Our workshop takes a different perspective, making figure skating as recognition target, and provides a fine-grained large-scale dataset to promote deep network learning for HAU. Thus, our workshop will bridge the needs of industry and research in academia, and may accelerate the process of these computer vision technologies being used in real applications.

% \section{Acknowledgement}
% This work was partially supported by the Natural Science Foundation of China No.62102039, the Fundamental Research Funds for the Central Universities No.500422813.

\bibliographystyle{unsrt}  
\bibliography{references}  %%% Remove comment to use the external .bib file (using bibtex).

\vspace*{3\baselineskip}

\begin{figwindownonum}[0,l,{\mbox{\includegraphics[width=3cm]{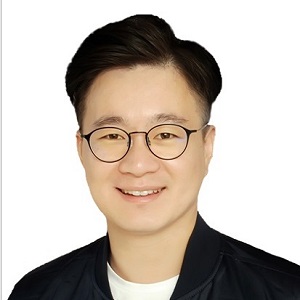}}},{}]
\noindent
\textbf{Jian Zhao} He is leader of Evolutionary Vision+x Oriented Learning (EVOL) Lab and Young Scientist at Institute of AI (TeleAI), China Telecom, and Researcher and Ph.D. Supervisor at Northwestern Polytechnical University (NWPU). He received his Ph.D. degree from National University of Singapore (NUS) in 2019 under the supervision of Assist. Prof. Jiashi Feng and Assoc. Prof. Shuicheng Yan. He is the SAC of VALSE, the committee member of CSIG-BVD, and the member of the board of directors of BSIG. He has received the "2020-2022 Youth Talent Promotion Project" from China Association for Science and Technology, and the "2021-2023 Beijing Youth Talent Promotion Project" from Beijing Association for Science and Technology. He has published over 40 cutting-edge papers on human-centric image understanding. He has won the Lee Hwee Kuan Award (Gold Award) on PREMIA 2019 and the "Best Student Paper Award" on ACM MM 2018 as the first author. He has received the nomination for the USERN Prize 2021, according to publications as first author in top rank (Q1) journals of the field of Artificial Intelligence, Pattern Recognition, Machine Learning, Computer Vision and Multimedia Analytics, in the recent two years. He has won the top-3 awards several times on world-wide competitions on face recognition, human parsing and pose estimation as the first author. His main research interests include deep learning, pattern recognition, computer vision and multimedia. He and his collaborators has also successfully organized the CVPR 2020 Anti-UAV Workshop and Challenge, the ECCV 2020 RLQ-TOD Workshop and Challenge, and the CVPR 2018 L.I.P Workshop and MHP Challenge.
\end{figwindownonum}
\vspace{0.5 cm}

\begin{figwindownonum}[0,l,{\mbox{\includegraphics[width=3cm]{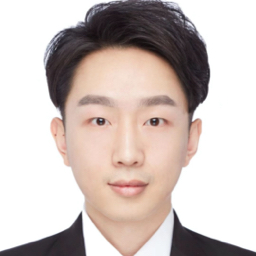}}},{}]
\noindent
\textbf{Lei Jin} is currently an Associate Research Fellow with the Beijing University of Posts and Telecommunications (BUPT), Beijing, China. He graduated from Beijing University of Posts and Telecommunications, his major research areas include computer vision, data mining, pattern recognition, with in-depth research in sub-fields such as human pose estimation, human action recognition, and human parsing, with related research results published in high-level conferences and journals such as CVPR, AAAI, NIPS, TMM, IJCAI, and ACMMM, and so on.
\end{figwindownonum}
\vspace{1 cm}

\begin{figwindownonum}[0,l,{\mbox{\includegraphics[width=3cm]{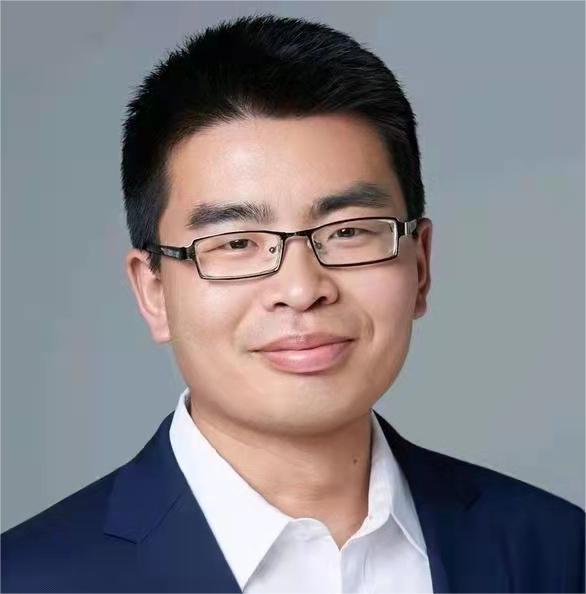}}},{}]
\noindent
\noindent\textbf{{Dr. Zheng Zhu}} is currently the Research Director and Scientist at PhiGent Robotics. 
During 2019-2021, he was a post-doc fellow at Tsinghua University, worked with Prof. Jiwen Lu. He received Ph.D. degree from Institute of Automation, Chinese Academy of Sciences in 2019. During 2016-2019, he was research interns at SenseTime, Horizon Robotics, and DeepGlint. He served reviewers in various journals and conference including IEEE-TPAMI, IEEE-TMM, IEEE-TCSVT, CVPR, ICCV, ECCV, ICLR. 
He has co-authored 40+ journal and conference papers mainly on computer vision and robotics problems, such as face recognition, visual tracking, human pose estimation, and servo control. 
He has more than 3,500 Google Scholar citations to his work, including SiamRPN (1,000+ citations) and DaSiamRPN (600+ citations). His work DaSiamRPN is included in famous OpenCV Library. 
He organized the Masked Face Recognition Challenge \& Workshop (MFR) in ICCV 2021. He ranked the 1st on NIST-FRVT Masked Face Recognition, won the COCO Keypoint Detection Challenge in ECCV 2020 and Visual Object Tracking (VOT) Real-Time Challenge in ECCV 2018.
\end{figwindownonum}
\vspace{1 cm}

\begin{figwindownonum}[0,l,{\mbox{\includegraphics[width=3cm]{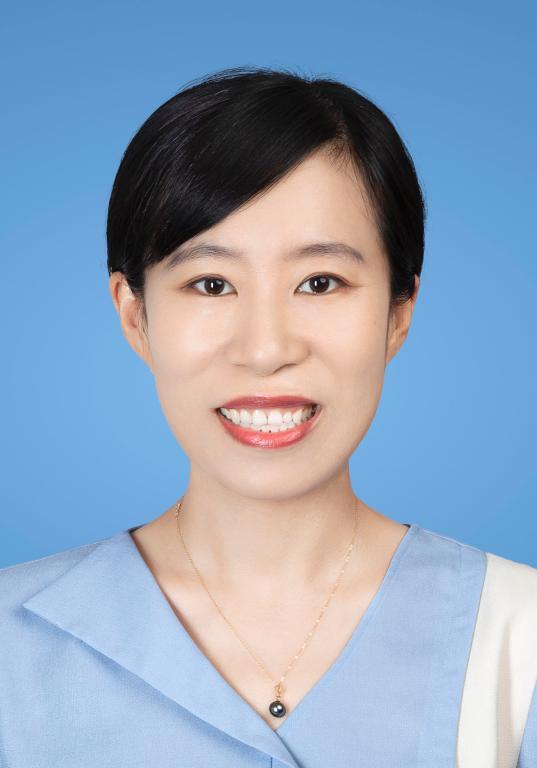}}},{}]
\noindent
\noindent\textbf{{Dr. Yinglei Teng}} is the professor of Beijing University of Posts and Telecommunication. 
She have won the National Natural Science Foundation of China, national key research and development, national nature fund instrument special, Huawei and other projects.
Currently, her scientific research interests include: edge computing and edge intelligence, synaesthesia, industrial Internet and so on. 
In the past five years, she have published 23 SCI papers (including 14 JRC 1), including one of the top authoritative journals in the field of communications and engineering, such as IEEE Communications Surveys \& Tutorials (IF:23.7), IEEE WCM (IF:11.391), IEEE TII (IF:9.112), IEEE TWC (IF:6.779), IEEE TVT (IF:5.379) and IEEE TCOM (IF:5.646). There are more than 50 papers published in CCF CCF B international conferences and EI journals such as IEEE infocom/globecom. Three published papers were listed as one of the top 1\% "highly cited papers" by ESI; one paper won the International Conference IEEE ICCAIS 2018 "Best thesis"; and she helped to guide the doctoral thesis to win the "excellent doctoral thesis of the Chinese Society of Electronic Education". Her authorized invention patent more than 50 projects, 1 CCSA industry standard (submitted for approval), 3 (established). 
She won the special prize of China Association for the Promotion of Science and Technology, the second prize of Beijing, and the first prize of science and technology of China Railway Society, etc.
\end{figwindownonum}
\vspace{0.1 cm}

\begin{figwindownonum}[0,l,{\mbox{\includegraphics[width=3cm]{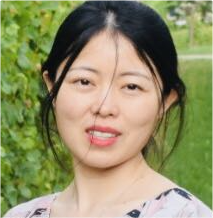}}},{}]
\noindent
\noindent\textbf{{Dr. Jiaojiao Zhao}} obtained her PhD degree in the VIS Lab, University of Amsterdam, the Netherlands. 
Before her PhD, she worked as a research assistant in the Artificial Intelligence Lab, University of East Anglia, UK. And from 2015 to 2016, she worked in Panasonic R\&D Center Singapore and visited the Learning and Vision Group at National University of Singapore. 
Her current research interests are video analysis including human action detection, tracking and recognition.
\end{figwindownonum}
\vspace{0.5 cm}

\begin{figwindownonum}[0,l,{\mbox{\includegraphics[width=3cm]{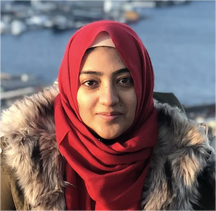}}},{}]
\noindent
\noindent\textbf{{Ms. Sadaf Gulshad}} is currently a Postdoc in the Multix Lab, University of Amsterdam, the Netherlands. 
Previously, she did her PhD in ML and CV from Bosch Delta Lab, University of Amsterdam. 
Before starting her PhD she completed her Masters in Electrical Engineering from KAIST, South Korea. 
Her current research interests lie in understanding and explaining the decisions of neural networks for video action classification. 
Previously she worked on robustifying neural networks against adversarial and natural perturbations.
\end{figwindownonum}
\vspace{0.5 cm}

\begin{figwindownonum}[0,l,{\mbox{\includegraphics[width=3cm]{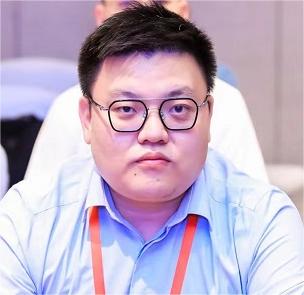}}},{}]
\noindent
\noindent\textbf{\underline{Dr. Zheng Wang}} is a Professor with Wuhan University, Wuhan, China. He was a Project Researcher and JSPS Fellowship Researcher with the National Institute of Informatics, Japan from 2017 to 2020, and a Project Assistant Professor with the Research Institute for an Inclusive Society through Engineering (RIISE), the University of Tokyo, in 2021. 
He had co-organized the ACM MM 2020 Tutorial on “Effective and Efficient: Toward Open-world Instance Re-identification”, CVPR 2020 Tutorial on “Image Retrieval in the Wild”, and ACM MM 2022 Tutorial on “Multimedia Content Understanding in Harsh Environments”. 
His research interests focus on multimedia content analysis.
\end{figwindownonum}
\vspace{0.5 cm}

\begin{figwindownonum}[0,l,{\mbox{\includegraphics[width=3cm]{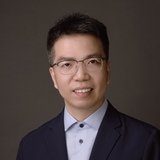}}},{}]
\noindent
\noindent\textbf{\underline{Dr. Bo Zhao}} is currently an applied AI researcher at BMO AI, based in Toronto, Canada. 
Prior to that, he worked at the University of British Columbia as a Postdoctoral Research Fellow with Prof. Leonid Sigal. 
He received his Ph.D. and B.Sc. degree from Southwest Jiaotong University, Chengdu, China. 
He also spent the last two years of his Ph.D. in the Learning and Vision Research Group at National University of Singapore as a visiting student with Prof. Jiashi Feng and Prof. Shuicheng Yan. 
His research interests include multimedia, computer vision, and machine learning. He has published several papers in top conferences and journals including CVPR, ECCV, MM, IJCV and TMM.
\end{figwindownonum}
\vspace{0.5 cm}

\begin{figwindownonum}[0,l,{\mbox{\includegraphics[width=3cm]{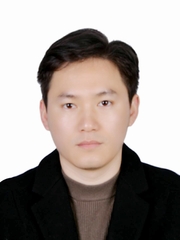}}},{}]
\noindent\textbf{\underline{Dr. Xiangbo Shu}} is currently a Professor at the School of Computer Science and Engineering, Nanjing University of Science and Technology, China. 
He received his Ph.D. degree from Nanjing University of Science and Technology, in 2016. From 2014 to 2015, he worked as a visiting scholar at the National University of Singapore, Singapore. 
His current research interests include Computer Vision, and Multimedia.
He has authored over 80 journal and conference papers in these areas, including IEEE TPAMI, IEEE TNNLS, IEEE TIP, CVPR, ICCV, ECCV, and ACM MM, etc. He has received the Best Student Paper Award in MMM 2016, and the Best Paper Runner-up in ACM MM 2015. 
He is the Senior Member of CCF, the Member of ACM, and the Senior Member of IEEE.
\end{figwindownonum}
\vspace{0.5 cm}

\begin{figwindownonum}[0,l,{\mbox{\includegraphics[width=3cm]{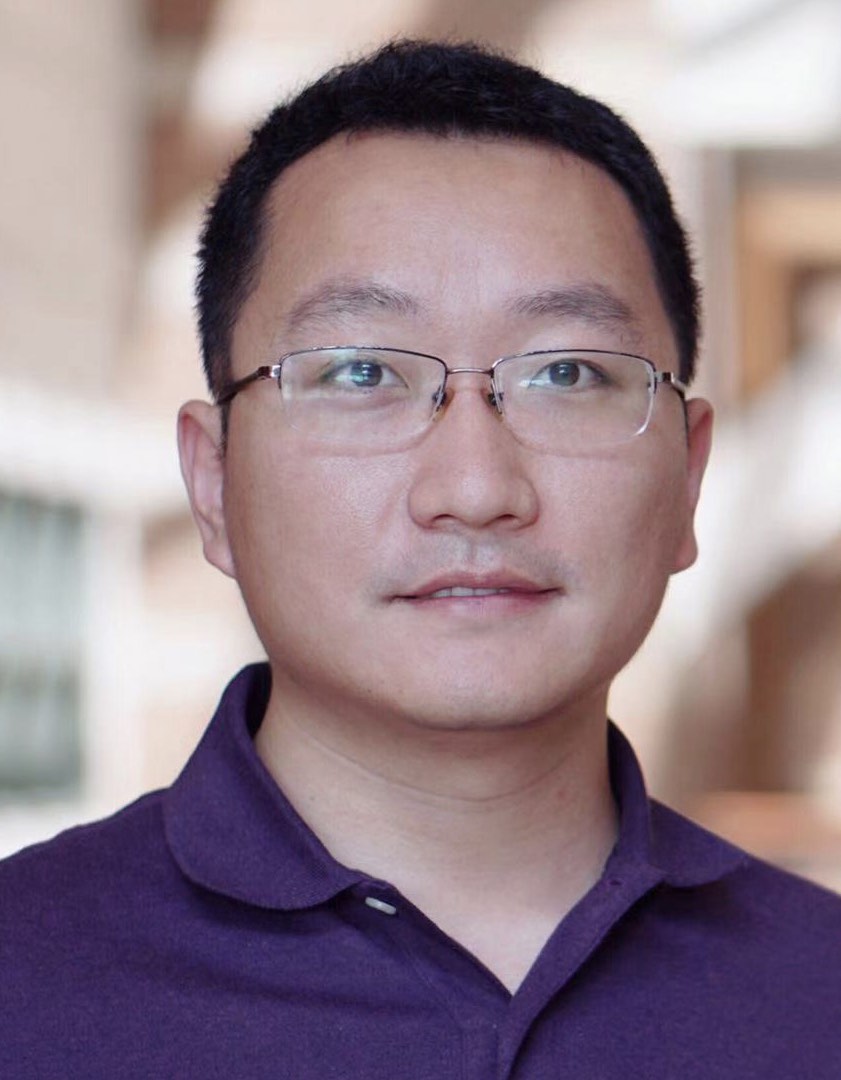}}},{}]
\noindent
\noindent\textbf{\underline{Dr. Yunchao Wei}} is currently a full professor of Beijing Jiaotong University. 
He was selected as MIT TR35 China by MIT Technology Review in 2021 and was named as one of the five top early-career researchers in Engineering and Computer Sciences in Australia by The Australian in 2020. He received the Discovery Early Career Researcher Award of the Australian Research Council in 2019, the 1st Prize in Science and Technology awarded by China Society of Image and Graphics (CSIG) in 2019. 
He has published more than 100 papers in top-tier conferences/journals, Google citations 13000+. He received many competition prizes from CVPR/ICCV/ECCV, such as the Winner prizes of ILSVRC 2014, LIP 2018/2019, Youtube VOS 2021, Runner-up Prizes of ILSVRC 2017, DAVIS 2020, etc. 
He organized many workshops on top-tier conferences, including Learning from Imperfect Data Workshop series (CVPR 2019, 2020, 2021) and Real-world Recognition from Low-quality Inputs Workshop series (ICCV 2019, ECCV 2020). He has broad research interests in computer vision and machine learning. 
His current research interest focuses on visual recognition with imperfect data, image/video segmentation and object detection, and multi-modal perception. 
\end{figwindownonum}
\vspace{0.5 cm}

\begin{figwindownonum}[0,l,{\mbox{\includegraphics[width=3cm]{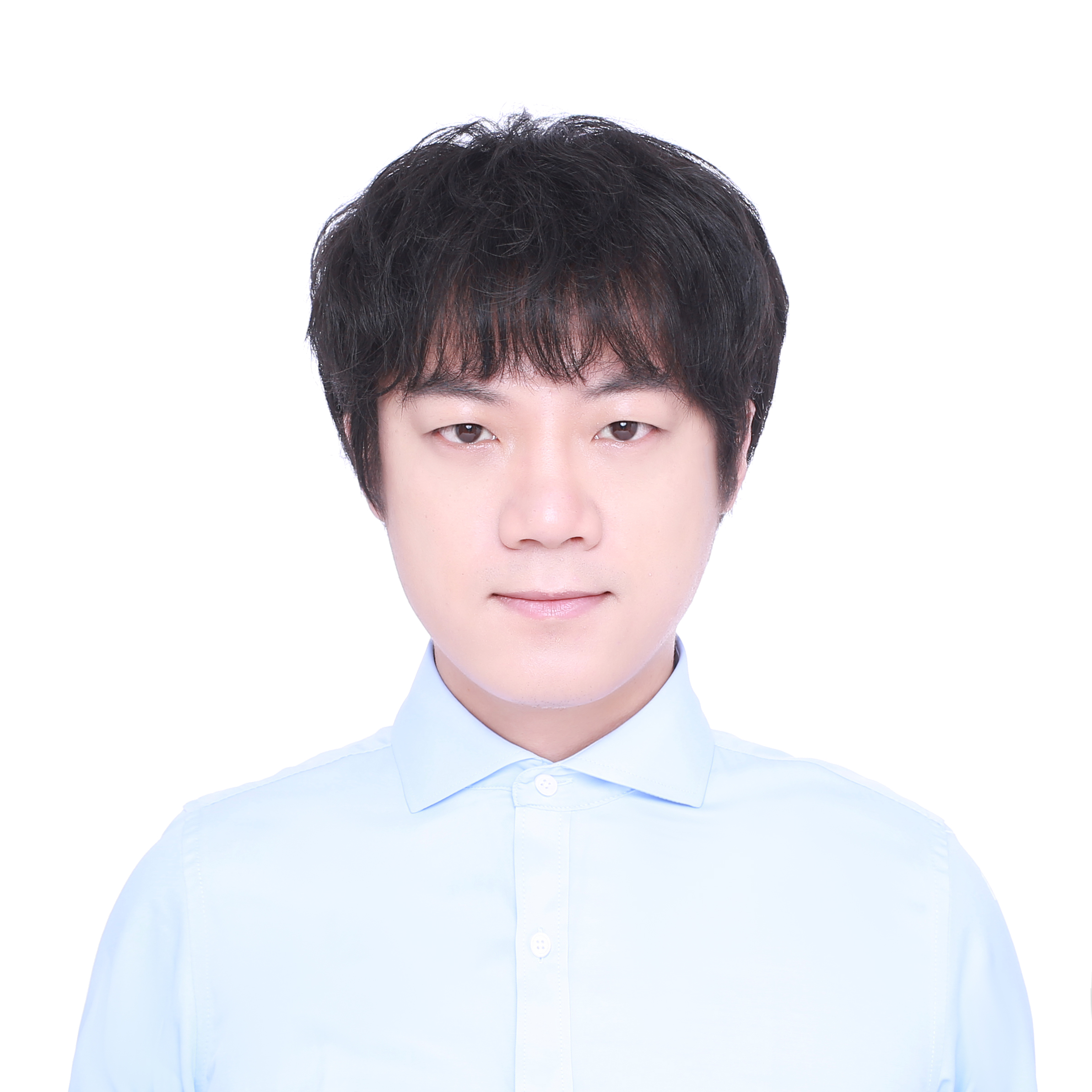}}},{}]
\noindent
\noindent\textbf{\underline{Dr. Xuecheng Nie}} received the B.S. and M.Eng. degrees in School of Computer Software from Tianjin University, Tianjin, China, in 2012 and 2015, respectively. He received the Ph.d. degree in Electrical and Computer Engineering from Nation University of Singapore, Singapore, in 2020. 
His research interests focus on Computer Vision, Deep Learning, Machine Learning, specially at Human Pose Estimation, Human Mesh Recovery and Action Recognition. He has served as the reviewers of T-PAMI, IJCV, T-MM, CVPR, ICCV, ECCV, NeurIPS, ICLR, LCML.
\end{figwindownonum}
\vspace{0.5 cm}

\begin{figwindownonum}[0,l,{\mbox{\includegraphics[width=3cm]{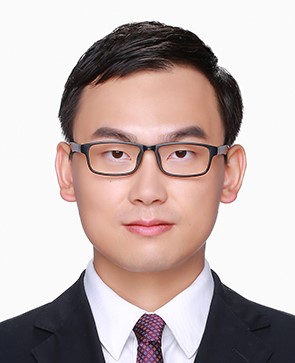}}},{}]
\noindent
\noindent\textbf{\underline{Dr. Xiaojie Jin}} is working as a Research Scientist in Bytedance Inc. in the USA. 
He received his Ph.D degree from National University of Singapore in 2018. 
His current research interests include self-supervised learning, multi-modal understanding, intelligent video editing and efficient deep model. His research works have been published in top-tier conferences/journals, including ICCV, ECCV, CVPR, ICML, NeurIPS, ICLR, TPAMI, TNNLS, etc.
\end{figwindownonum}
\vspace{1.5 cm}

\begin{figwindownonum}[0,l,{\mbox{\includegraphics[width=3cm]{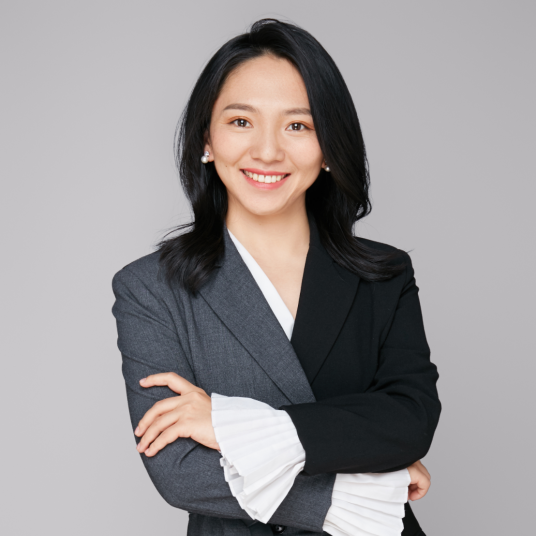}}},{}]
\noindent
\noindent\textbf{\underline{Dr. Xiaodan Liang}} is currently an Associate Professor at Sun Yat-sen University. 
She was a Project Scientist at Carnegie Mellon University, working with Prof. Eric Xing. 
She focuses on interpretable and cognitive intelligence and its applications on large-scale visual recognition, automatic machine learning and cross-modality dialogue systems. 
She serves as an Area Chair of ICCV 2019, CVPR 2020 and Tutorial Chair (Organization committee) of CVPR 2021. She has been awarded ACL 2019 Best Demo paper nomination. 
She and her collaborators has also published the largest human parsing dataset. And she successfully organized workshops and challenges on CVPR 2017, CVPR 2018, CVPR 2019, CVPR 2020, ICML 2019 and ICLR 2021.
\end{figwindownonum}
\vspace{0.5 cm}

\begin{figwindownonum}[0,l,{\mbox{\includegraphics[width=3cm]{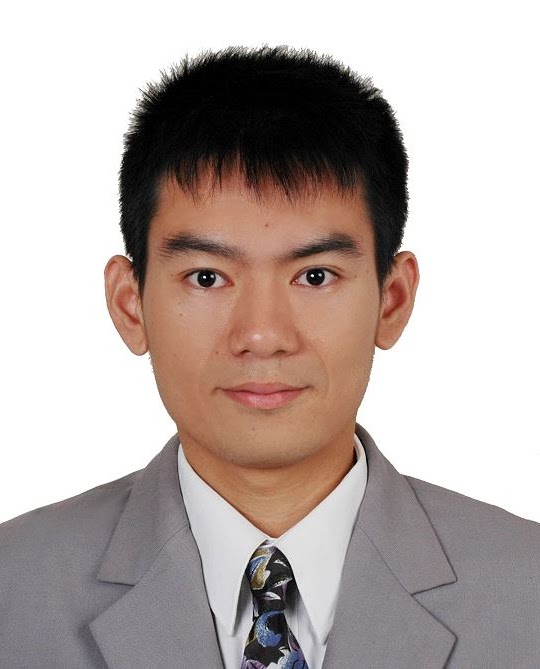}}},{}]
\noindent
\noindent\textbf{\underline{Dr. Jianshu Li}} 
obtained the Ph.D. degree from School of Computing, National University of Singapore in 2019, advised by Prof. Terence Sim and Prof. Shuicheng Yan. 
His research interest is mainly computer vision and image understanding, particularly face and human analytics, semantic segmentation and object detection. 
has been working as an algorithm expert in Ant Group since 2018, mainly working on face analysis algorithms, including face recognition, face liveness detection, face quality analysis and Deepfake detection, etc. 
He is the winner of the gold award of PREMIA 2019 Singapore, best student paper award of ACMMM 2018, winner prize of object localization ILSVRC 2017, winner prize of emotion recognition challenge ICMI 2016. 
He has published 20+ papers in journals and conferences and served as invited reviewers in CVPR, ECCV, NIPS, IJCAI, FG, ICMI, TIP, TCSVT, TMM etc. 
\end{figwindownonum}
\vspace{0.5 cm}

\begin{figwindownonum}[0,l,{\mbox{\includegraphics[width=3cm]{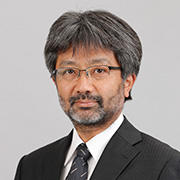}}},{}]
\noindent
\noindent\textbf{{Dr. Shin'ichi Satoh}} is a Professor at National Institute of Informatics (NII) and University of Tokyo. Satoh is currently the Director of Research Center for Medical Big Data, NII. 
He is a Fellow of IEICE and ITE and has served as EditorialBoard/Associate Editor for lots of top Journals, such as IJCV, TMM, TCSVT. He has published 4 graduate level textbooks and over 250 scientific papers in well recognized journals including TIP, TMM, TCSVT, TCYB, and top international conference including ACM MM, CVPR, ICCV, ECCV, AAAI, IJCAI. 
Satoh’s research interests include multimedia information analysis and knowledge discovery, aiming to create an intelligent computer system to see and understand the visual world.
\end{figwindownonum}

\begin{figwindownonum}[0,l,{\mbox{\includegraphics[width=3cm]{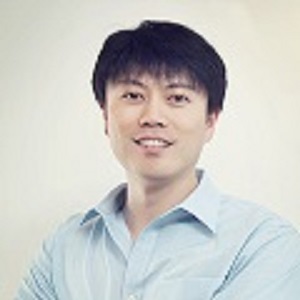}}},{}]
\noindent
\textbf{Yandong Guo}  is currently the OPPO Chief Scientist of Intelligent Perception and 
Adjunct Professor of Beijing University of Posts and Telecommunications. He gained his 
PhD from School of Electronic and Computer Engineering, Purdue University (West 
Lafayette) under the supervision of Prof. Jan P. Allebach and Charles A. Bouman. Dr. Guo 
mainly focuses on computer vision and artificial intelligence research and applies these 
research in industry applications. His papers have been widely accepted in CVPR, ECCV, 
ICML and other internationally recognized academic conferences and journals, and have 
been cited thousands of times by peers, and have also empowered many core products of 
companies including GE, HP, Microsoft, Xpeng Motors and OPPO. He has also served as a 
program committee member or reviewer for multiple conferences and journals and has 
organized ICCV and CVPR Workshops as the chair member.
\end{figwindownonum}

\begin{figwindownonum}[0,l,{\mbox{\includegraphics[width=3cm]{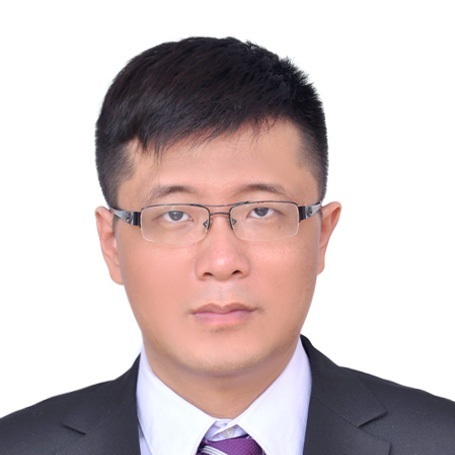}}},{}]
\noindent
\noindent\textbf{{Dr. Cewu Lu}}  is a Professor at Shanghai Jiao Tong University (SJTU). His research interests fall mainly in Computer Vision and Embodied AI. The well-known projects are Aphapose, HAKE, GraspNet. He served as Area Chair of CVPR 2020/2021/2022 and ICCV 2021, ECCV 2022.
\end{figwindownonum}
\vspace{15mm}

\begin{figwindownonum}[0,l,{\mbox{\includegraphics[width=3cm]{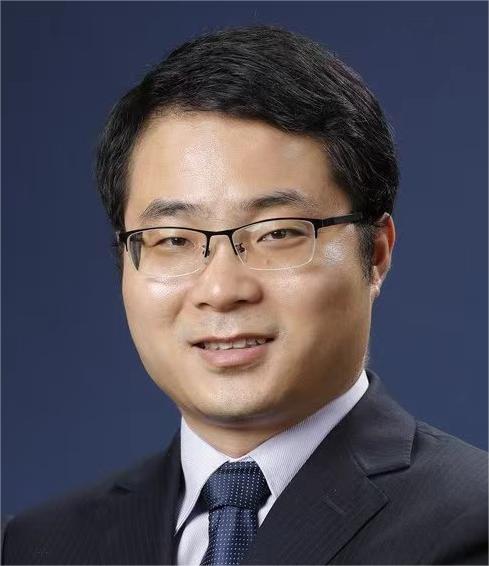}}},{}]
\noindent
\noindent\textbf{{Dr. Junliang Xing}} is a Professor with Tsinghua University and the recipient of the National Science Fund for Excellent Young Scholar. 
He obtained his dual bachelor's degrees in Computer Science and Mathematics at Xi'an Jiaotong University in 2007 and his doctorate in Computer Science in 2012. Then he worked in the National Laboratory of Pattern Recognition, Institute of Automation, Chinese Academy of Sciences as an assistant researcher, associated researcher, and researcher in 2012, 2015, and 2018, respectively. 
His research interests are human-computer interactive learning, computer gaming, and computer vision. He has published more than 100 peer-reviewed papers in international conferences and journals and got more than 13,000 citations from Google Scholar. 
He has also published two books and translated three classical textbooks in artificial intelligence. He has been granted the "New People in Academia", "Google Fellowship", and three times "Best/Distinguished Paper" awards in top international and domestic conferences, with a dozen times of prizes in AI technique competitions. His research techniques have got several applications in practical systems.
\end{figwindownonum}

\begin{figwindownonum}[0,l,{\mbox{\includegraphics[width=3cm]{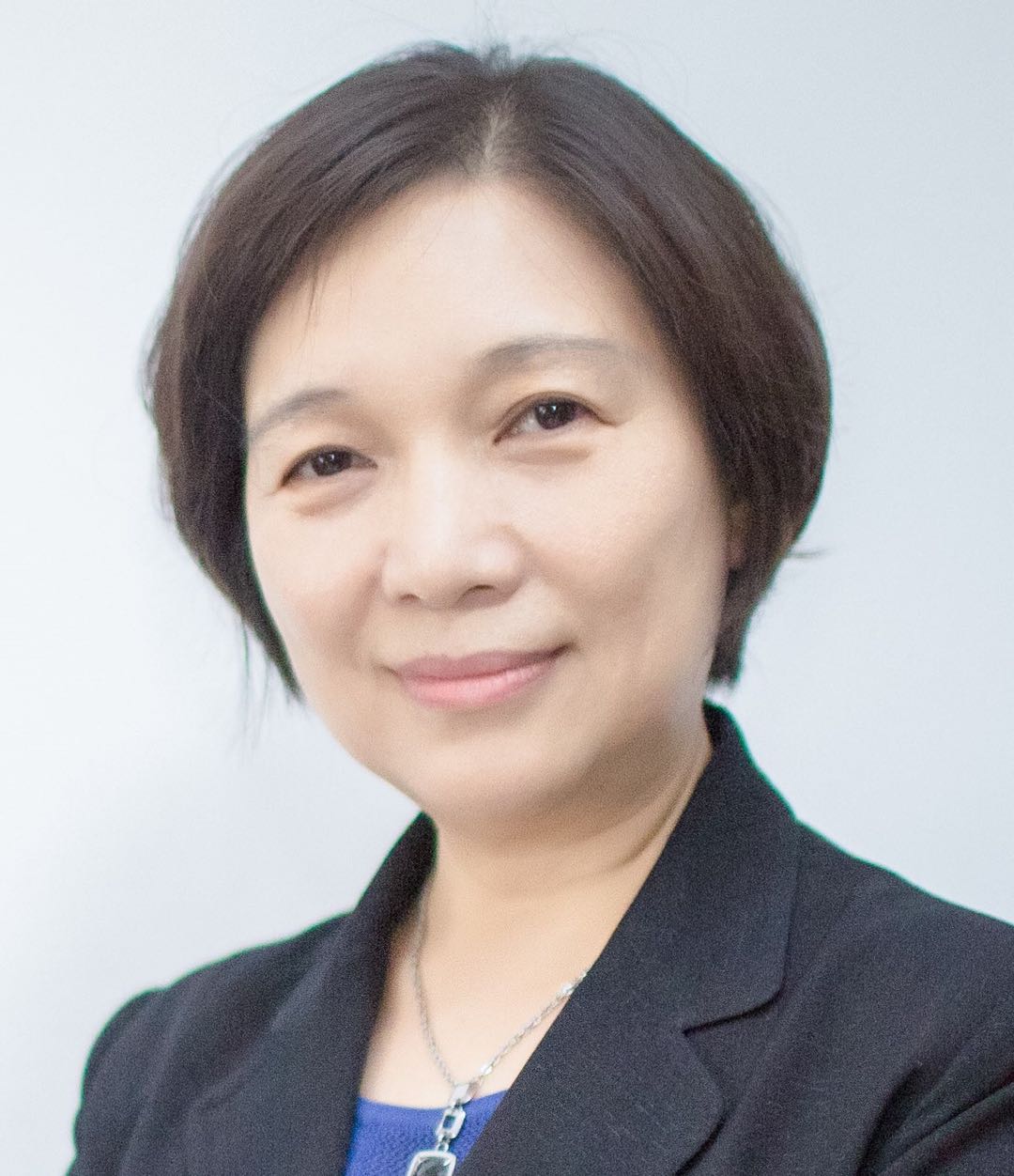}}},{}]
\noindent
\noindent\textbf{{Dr. Jane Shen Shengmei}} is the Chief Scientist of Mikomiko, and she is specialized in AI, Deep Learning, Face \& Image Recognition, 3D, Autonomous Driving, Image/video/audio Processing and Compression. Deep experience in leading research and technology teams in computer vision, AI and robotics domains, with highly cited publications in top journals/conferences and exceptional accomplishments in international competitions. 
Published and led research for over 150 papers and patents, with publications in venues including CVPR, NeurIPS, ICCV, ECCV, AAAI, CoRL, ICIP, ICPR, and Google Scholar profile of over 4800 citations, h-index of 38 and i10 index of 91. 
Directed team research and provided hands-on technical expertise for computer vision algorithm design that resulted in 1st place results in PASCAL VOC 2010, 2011, 2012 PASCAL VOC, 1st place result in Visual Object Tracking in 2013, 1st place result in Microsoft 1M-Celebrity facial recognition competition 2017 (track 1 and 2), 1st place in anomaly detection track for CVPR 2018 AI City Challenge, 1st place for IROS 2018 mobile robotics challenge, 1st place for CVPR 2019 lightweight facial recognition challenge across all 3 tracks. 
Research work directly translated into industry applications through Panasonic/Pensees products and solutions. Recognizedthought leader in the Asia-Pacific region and in Singapore for industry contributions in computer vision, AI and machine learning products. Awarded inaugural 100 Women in Tech Award 2020 by Singapore government, IT Awards Leader 2021 in Entrepreneurship by Singapore Computer Society.
\end{figwindownonum}

\end{document}